\documentclass[runningheads]{llncs}
\usepackage{graphicx}
\usepackage{comment}
\usepackage{amsmath,amssymb} 
\usepackage{color}
\usepackage{subcaption}
\usepackage{enumitem}
\usepackage[amssymb]{SIunits}
\usepackage{booktabs}
\usepackage{seqsplit}
\usepackage{threeparttable}
\usepackage{afterpage}
\usepackage{nicefrac}
\usepackage{multirow}
\usepackage{paralist}
\usepackage{wrapfig}
\usepackage{hyperref}

\let\llncssubparagraph\subparagraph
\let\subparagraph\paragraph
\usepackage[compact]{titlesec}
\let\subparagraph\llncssubparagraph
\titlespacing{\section}{0pt}{2ex}{1ex}
\titlespacing{\subsection}{0pt}{1ex}{0.2ex}
\titlespacing{\subsubsection}{0pt}{0.3ex}{0.1ex}

\makeatletter
\newcommand*{\rom}[1]{\expandafter\@slowromancap\romannumeral #1@}

\DeclareMathOperator{\softmax}{softmax}

\newcommand{\etal}{\textit{et al.}~}
\newcommand{\ie}{\textit{i.e.},~}
\newcommand{\eg}{\textit{e.g.},~}
\newcommand{\etc}{\textit{etc.}~}
\newcommand{\fref}[1]{Fig.~\ref{#1}}
\newcommand{\sref}[1]{Sec.~\ref{#1}}
\newcommand{\tref}[1]{Table~\ref{#1}}

\newcommand{\mysec}[1]{\noindent\textbf{#1}}
\newcommand{\mysubsec}[1]{\textit{#1}}

\usepackage{array}
\newcommand{\PreserveBackslash}[1]{\let\temp=\\#1\let\\=\temp}
\newcolumntype{C}[1]{>{\PreserveBackslash\centering}p{#1}}
\newcolumntype{R}[1]{>{\PreserveBackslash\raggedleft}p{#1}}
\newcolumntype{L}[1]{>{\PreserveBackslash\raggedright}p{#1}}

\newcommand{\orcid}[1]{\href{https://orcid.org/#1}{\includegraphics[width=8pt]{icons/ORCID-iD_icon-64x64.png}}}

\begin{document}
\pagestyle{headings}
\mainmatter
\def\ECCVSubNumber{2904}  

\title{Visual Memorability for Robotic Interestingness\\via Unsupervised Online Learning} 

\titlerunning{Unsupervised Online Learning for Visual Interestingness}
\author{Chen Wang\orcidID{0000-0002-4630-0805}\and\\
Wenshan Wang\orcidID{0000-0002-4488-5619}\and
Yuheng Qiu\orcidID{0000-0003-2966-8103}\and\\
Yafei Hu\orcidID{0000-0002-8142-1629}\and
Sebastian Scherer\orcidID{0000-0002-8373-4688}
}
\authorrunning{Chen Wang et al.}
\institute{Carnegie Mellon University, Pittsburgh, PA 15213, USA\\
\email{chenwang@dr.com, \{wenshanw,yuhengq,yafeih,basti\}@andrew.cmu.edu}\\
\url{https://github.com/wang-chen/interestingness}
}
\maketitle

\begin{abstract}
In this paper, we explore the problem of interesting scene prediction for mobile robots.
This area is currently underexplored but is crucial for many practical applications such as autonomous exploration and decision making.
Inspired by industrial demands, we first propose a novel translation-invariant visual memory for recalling and identifying interesting scenes, then design a three-stage architecture of long-term, short-term, and online learning.
This enables our system to learn human-like experience, environmental knowledge, and online adaption, respectively.
Our approach achieves much higher accuracy than the state-of-the-art algorithms on challenging robotic interestingness datasets.

\end{abstract}

\section{Introduction}\label{sec:introduction}

\begin{figure*}[!t]
    \centering
    \includegraphics[width=1.0\textwidth]{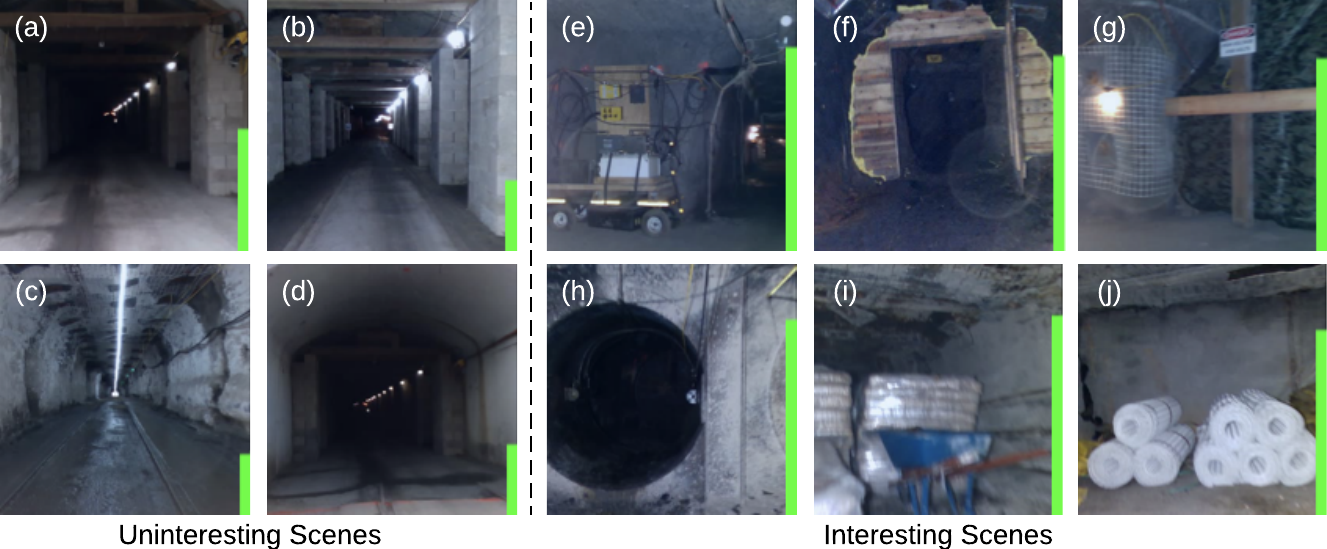}
    \caption{In this paper, we aim to predict robotic interesting scenes, which are crucial for decision making and autonomous cooperation. To enable the behavior of online losing interests on repetitive scenes for exploration of mobile robots, we propose to establish an online update scheme for interesting scene prediction. This figure shows several examples of both uninteresting and interesting scenes in SubT data \cite{subtdata} taken by autonomous robots. The height of green strip located at the right of each image indicates the interestingness level predicted by our unsupervised online learning algorithm when it sees the scene for the first time.
    }
    \label{fig:motivation}
\end{figure*}

Interesting scene prediction is crucial for autonomous exploration \cite{osswald2016speeding}, which is one of the most fundamental capabilities of mobile robots.
It has a significant impact on decision making and robot cooperation.
For example, the finding of a door shown in \fref{fig:motivation} (f) may affect the future planing, the hole on the wall in \fref{fig:motivation} (h) may attract more attentions.
However, prior algorithms often have difficulty when they are deployed to unknown environments, as the robots not only have to find interesting scenes, but also have to lose the interests on repetitive scenes, \ie interesting scenes may become uninteresting during robot exploration after repeatedly observing similar scenes or following moving objects.
For example in \fref{fig:drone-filming}, we expect to have high interests on the truck when it appears but loss the interests when it exists for a long time.
Nevertheless, the recent approaches of interestingness detection \cite{gygli2016analyzing,jiang2013understanding}, as well as saliency detection \cite{zhang2017learning}, anomaly detection \cite{zhao2011online,luo2017revisit}, novelty detection \cite{abati2019latent}, and meaningfulness detection \cite{hasan2016learning} algorithms cannot achieve this online updates scheme.

To this end, we propose to establish an \textbf{online learning} scheme to search for interesting scenes for robot exploration tasks.
On the other hand, existing algorithms are heavily dependent on back-propagation algorithm \cite{rumelhart1988learning} for learning, which is very computational expensive.
To solve this problem, we introduce a novel \textit{translation-invariant 4-D visual memory} to identify and recall visually interesting scenes.
Human beings have a great capacity to direct visual attention and judge the interestingness of a scene \cite{constantin2019computational}.
For mobile robots, we find the following properties are necessary to establish a sense of visual interestingness.

\subsubsection{Unsupervised:}~
As shown in \fref{fig:motivation}, the interesting scenes in robot operating environments are often unique and unknown, thus the labels are normally difficult to obtain, but prior research mainly focuses on supervised methods \cite{amengual2015review,constantin2019computational} and suffers in prior unseen environments.
We hypothesize that a sense of interestingness can be established in an unsupervised manner.

\subsubsection{Task-dependent:}~
In many practical applications, we might only know uninteresting scenes before a mission is started. In the example of tunnel exploration task in \fref{fig:motivation}, the deployment will be more efficient and easier if the robots can be taught what is uninteresting within several minutes. In this sense, we argue that the visual interestingness prediction system should be able to learn from negative samples quickly without accessing to data from unsupervised learning, thus an incremental learning method is necessary. Note that we expect the model is capable of learning from negative samples, but it is not necessary for all tasks.

To achieve the above properties, we propose a three-stage architecture:

\subsubsection{Long-term learning:}~
In this stage, we expect a model to be trained off-line on a large amount of data in an unsupervised manner as human beings acquire common knowledge from experience. We also expect the training time on single machine to be no more than the order of days.

\subsubsection{Short-term learning:}~For task-dependent knowledge, the model then should be able to learn from hundreds of uninteresting images in minutes. This can be done before a mission started and beneficial to quick robot deployment.

\subsubsection{Online learning:}~During mission execution the system should express the top interests in real-time and the detected interests should be lost online when similar scenes appear frequently, regardless if they exist in the uninteresting images or not.
Another important aspect for online learning is no data leakage, \ie each frame is proceed without using information from its subsequent frames. This is in contrast to prior works \cite{jiang2013understanding,gygli2016analyzing} and datasets \cite{demarty2017predicting}, where interesting frames are selected after an entire sequence is processed \cite{grabner2013visual}. Since robots need to respond in real-time, we require that our algorithms are able to adapt quickly. To measure such capability of online response, we propose a new evaluation metric.

In summary, our contributions are:
\begin{compactitem}
    \item We introduce an extremely simplified three-stage architecture for robotic interesting scene prediction, which is crucial for practical applications. Concretely, we leverage long-term learning to acquire human-link experience, short-term learning for quick robot deployment and task-related knowledge, and online learning for environment adaption and real-time response.
    \item To accelerate the short-term and online learning, we propose a novel 4-D visual memory to replace back-propagation. Concretely, we introduce cross-correlation similarity for translational invariance, which is crucial for perceiving video stream, we also introduce tangent operator for safe writing, which is crucial for incremental learning from negative samples.
    \item To measure the online performance, we propose a strict evaluation metric, \ie the area under the curve of online precision (AUC-OP) to jointly consider precision, recall rate, and online performance.
    \item It is demonstrated that our approach achieves much higher overall performance than the state-of-the-art algorithms.
\end{compactitem}

\section{Related Work}\label{sec:related-work}

A learning system that encodes the three-stage architecture for interesting scene prediction has not been achieved, thus the formulation as well as performance evaluation will be quite different from prior approaches.
Some works on interestingness prediction have different objectives \cite{amengual2015review}, \eg Shen \etal aimed to predict human interestingness on social media \cite{shen2017deep}. In this section we will mainly review the related techniques, as some methods used in saliency, anomaly, and novelty detection are also useful for our work.

The definition of interestingness is subjective, thus the annotation has to be averaged over different participants.
To mimic the human judgment, prior works have paid great attentions to investigate the relationship between human visual interestingness and image features \cite{amengual2015review}.
They are typically inspired by psychological cues and heavily leverage human annotation for training, which results in a large family of supervised learning methods.
For instance, Dhar \etal designed three hand-crafted rules, including attributes of composition, content, and sky-illumination to approximate both aesthetics and interestingness of images \cite{dhar2011high}.
Jiang \etal extended image interestingness to video and evaluated hand-crafted visual features for predicting interestingness on the YouTube and Flickr datasets \cite{jiang2013understanding}.
Fu \etal formulated interestingness as a problem of unified learning to rank, which is able to jointly identify human annotation outliers \cite{fu2014interestingness,fu2015robust}.

Deep neural networks played more and more significant roles in recent works on interestingness prediction. For example, Gygli \etal introduced VGG features \cite{Simonyan:2015ws} and leveraged a support vector regression model to predict the interestingness of animated GIFs \cite{gygli2016analyzing}.
Chaabouni \etal constructed a customized CNN model to identify salient and non-salient windows for video interestingness prediction \cite{chaabouni2017deep}.
Inspired by a human annotation procedure of pairwise comparison, Wang \etal combined two deep ranking networks \cite{wang2018video} to obtain better performance, and this method ranked first in the 2017 interestingness prediction competition \cite{demarty2017mediaeval}.
Shen \etal combined both CNN and LSTM \cite{hochreiter1997long} for feature learning to predict video interestingness \cite{shen2017deep} for media contents.

However, the aforementioned methods are highly dependent on human annotation for training, which is labor expensive and not suitable for interestingness search \cite{constantin2019computational}.
Some efforts for unsupervised learning of interestigness have been made in \cite{ito2012detecting}, where interesting events of videos are detected using the density ratio estimation algorithm with the HOG feature \cite{dalal2005histograms}. However, in practice the approach cannot adapt well to changing distributions.

In the long-term stage, we introduce an autoencoder \cite{kramer1991nonlinear} for unsupervised learning, which has been widely used for feature extraction in many applications. For example, Hasan \etal  showed that an autoencoder is able to learn regular dynamics and identify irregularity in long-duration videos \cite{hasan2016learning}. Zhang \etal introduced dropout into the autoencoder for pixel-wise saliency detection in images \cite{zhang2017learning}.
Zhao \etal proposed a spatio-temporal autoencoder to extract both spatial and temporal features for anomaly detection \cite{zhao2017spatio}.

In order to learn online, we introduce a novel visual memory module into the convolutional neural networks.
Visual memory has been widely investigated in neuroscience \cite{phillips1974distinction}.
While in computer vision, memory aided neural networks received limited attentions and used for several different tasks.
For example, Graves et al. proposed a differentiable neural Turning machines (NTM) \cite{graves2014neural}, which coupled external memory with recurrent neural networks (RNN).
Santoro et al. extended NTM and designed a module to efficiently access the memory \cite{santoro2016meta}.
Gong et al. introduced memory module into an auto-encoder to remember normal events for anomaly detection \cite{gong2019memorizing}.
Kim et al. introduced the memory network into GANs to remember previously generated samples to alleviate the forgetting problem \cite{kim2018memorization}.
However, the memories in the above works are defined as flattened vectors, thus the spatial structural information cannot be retained. In this paper, we propose a translation-invariant memory module and introduce online learning to solve the problem of robotic interestingness prediction.

\section{Visual Memory}\label{sec:memory}

To retain the structural information of visual inputs, the visual memory $\mathbf{M}$ is defined as a 4-D tensor, \ie $\mathbf{M}\in\mathbb{R}^{n\times c\times h\times w}$, where $n$ is the number of memory cubes and $c$, $h$, and $w$ are the channel, height, and width of each cube, respectively.
Intuitively, memory writing is to encode visual inputs into the memory, while reading is to recall one's memory regarding the visual inputs.

\subsection{Memory Writing}

We desire that the visual memory is able to balance new visual inputs and old knowledge.
To this end, we denote visual inputs at time $t$ as $\mathbf{x}(t)\in\mathbb{R}^{c\times h\times w}$ and define the writing protocol for the i$_{th}$ memory cube $\mathbf{M}_i$ at time $t$ as
\begin{equation}\label{eq:writing}
\mathbf{M}_i(t) = (1-\mathbf{w}_i)\cdot \mathbf{M}_i(t-1) + \mathbf{w}_i\cdot \mathbf{x}(t),
\end{equation}
where $\mathbf{w}_i$ is the i$_{th}$ element of a weight vector $\mathbf{w}\in\mathbb{R}^n$,
\begin{equation}\label{eq:writing-vector}
\mathbf{w} = \sigma(\gamma_w\cdot\tan(\frac{\pi}{2}\cdot D(\mathbf{x}(t), \mathbf{M}(t-1)))),
\end{equation}
where $\sigma(\cdot)$ is the softmax function and $D(\mathbf{x}, \mathbf{M})$ is a cosine similarity vector, in which the i$_{th}$ element $D_i(\mathbf{x}, \mathbf{M})$ is
\begin{equation}\label{eq:cosine}
D_i(\mathbf{x}, \mathbf{M}) = \frac{\sum(\mathbf{x}\odot \mathbf{M}_i)}{\|\mathbf{x}\|_\mathbf{F}\cdot\|\mathbf{M}_i\|_\mathbf{F}},
\end{equation}
where $\odot$, $\sum$, and $\|\cdot\|_\mathbf{F}$ are element-wise product, elements summation, and Frobenius norm, respectively.
The writing protocol in \eqref{eq:writing} is a moving average, whose learning speed can be controlled via the writing rate $\gamma_w~(\gamma_w>0)$, so that the training samples can be learned with an expected speed.

It is worth noting that, to promote the sparsity of memory writing, we introduce a tangent operator in \eqref{eq:writing-vector} to map the range of cosine similarity $[-1,1]$ in \eqref{eq:cosine} to $[-\inf,\inf]$, thus memory writing can be focused on fewer but more relevant cubes via the softmax function.
This leads to easier incremental learning and efficient space usage, which will be further explained in \sref{sec:short-term} and \sref{sec:writing-vector}.

\subsection{Memory Reading}\label{sec:memory-reading}
Recall that convolutional features (visual inputs) are invariant to small input translations due to the concatenation of pooling layers to convolutional layers \cite{goodfellow2016deep}.
To obtain invariance to large translations, we need other techniques such as data augmentation, which is very computationally heavy.
To solve this problem, we introduce translation in memory reading, leveraging that the structural information of visual inputs are retained in memory writing.
Denote 2-D circular translation along the width and height directions with $(x,y)$ elements of the i$_{th}$ memory cube at time $t$ as $\mathbf{M}_i^{(x,y)}(t)$, memory reading $\mathbf{f}(t)\in\mathbb{R}^{c\times h \times w}$ is
\begin{equation}\label{eq:reading}
\mathbf{f}(t) = \sum_{i=1}^{n}\mathbf{r}_i\cdot \mathbf{M}_i^{(x,y)}(t),
\end{equation}
where $\mathbf{r}_i$ is the i$_{th}$ element of reading weight vector $\mathbf{r}\in \mathbb{R}^n$,
\begin{equation}\label{eq:reading-vector}
\mathbf{r} = \sigma(\gamma_r\cdot\tan(\frac{\pi}{2}\cdot S(\mathbf{x}(t), \mathbf{M}(t)))),
\end{equation}
where $\gamma_r>0$ is the reading rate.  The i$_{th}$ element of $S(\mathbf{x}, \mathbf{M})$ is the maximum cosine similarity of $\mathbf{x}$ with $\mathbf{M}_i^{(a,b)}$, where $a=0:h-1$ and $b=0:w-1$ imply all translations.
Intuitively, to find the maximum cosine similarity, we need to repeatedly compute \eqref{eq:cosine} for translated memory cube $h\times w$ times, resulting in a high computational complexity.
To solve this problem, we leverage the fast Fourier transform (FFT) to compute the cross-correlation \cite{wang2019kernel}.
Recall that 2-D cross-correlation is the inner-products between the first signal and circular translations of the second signal \cite{wang2018kernel}, we can compute $S_i(\mathbf{x}, \mathbf{M}_i)$ as
\begin{equation}\label{eq:cross-correlation}
S_i(\mathbf{x}, \mathbf{M}) = \frac{\max\mathcal{F}^{-1}(\sum^{c}\hat{\mathbf{x}}^*\odot \hat{\mathbf{M}}_i)}{\|\mathbf{x}\|_\mathbf{F}\cdot\|\mathbf{M}_i\|_\mathbf{F}},
\end{equation}
where $\hat{\cdot}$ is the 2-D FFT, $\cdot^*$ is the complex conjugate, and $\sum^{c}$ is element-wise summation along channel dimension.
The translation $(x,y)$ in \eqref{eq:reading} for the i$_{th}$ memory cube is corresponding to the location of the maximum response, \ie
\begin{equation}
(x,y) = \arg\max _{(a,b)} (\sum^{C}\hat{\mathbf{x}}^*\odot \hat{\mathbf{M}}_i)[a,b].
\end{equation}
In this way, the computational complexity for each memory cube can be reduced from $\mathcal{O}(ch^2w^2)$ to $\mathcal{O}(chw\log hw)$.
Another advantage of translation-invariance in memory reading is that memory usage becomes more efficient, since scene translation is common in video stream for many robotic applications, \eg robot exploration and object search, which will be further explained in
\sref{sec:translation-invariance}.

\section{Learning}\label{sec:learning}

\subsection{Long-term Learning}\label{sec:long-term}

\begin{figure*}[!t]
    \centering
    \begin{subfigure}[t]{0.32\textwidth}
        \centering
        \includegraphics[height=0.6\linewidth]{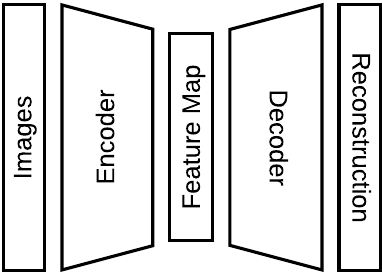}
        \caption{Long-term learning.}
        \label{fig:long-term-learning}
    \end{subfigure}
    \begin{subfigure}[t]{0.32\textwidth}
        \centering
        \includegraphics[height=0.6\linewidth]{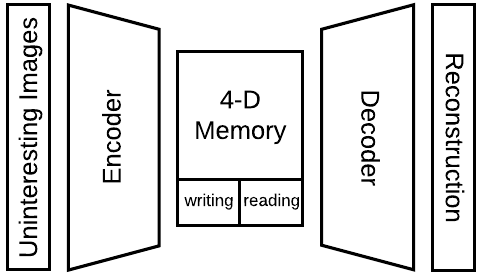}
        \caption{Short-term learning. }
        \label{fig:short-term-learning}
    \end{subfigure}%
    \hspace{0.02\textwidth}
    \begin{subfigure}[t]{0.32\textwidth}
        \centering
        \includegraphics[height=0.6\linewidth]{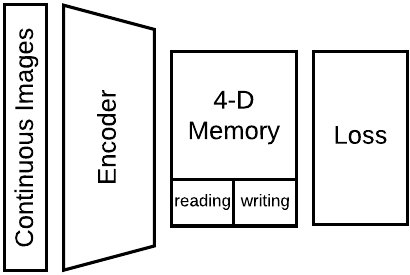}
        \caption{Online learning.}
        \label{fig:online-learning}
    \end{subfigure}
    \caption{The proposed three learning stages. (a) In long-term learning, the parameters in both encoder and decoder are trainable. (b) In short-term learning, the parameters in the encoder and decoder are frozen; the memory writing is performed before reading. (c) In online learning, the parameters in the encoder are frozen; the memory reading is performed before writing.}
    \label{fig:learning}
\end{figure*}

Inspired by the fact that human has a massive memory storage capacity \cite{brady2008visual}, we use an autoencoder in \fref{fig:long-term-learning} for long-term learning for the following reasons.

\mysec{Unsupervised Knowledge:} A reconstruction model can be trained in an unsupervised way, hence we can collect massive number of images from the internet or in real-time during execution to train the model without much efforts.
This agrees with our objective of long-term learning that is to remember as many scenes as possible.
In this stage, we still leverage back-propagation for the training, so that the large amount of knowledge will be `stored' in the trainable parameters, which will be frozen afterwards. In this sense, the learned knowledge can be treated as unforgettable human-like experience.

\mysec{Detailed and Semantic:} To precisely reconstruct images by smaller feature maps, the output of the bottleneck layer has to contain both detailed and semantic information.
This is crucial for visual interestingness, since both texture and object-level information may attract one's interests.
Feature maps are invariant to small translations due to CNN, we leverage the invariance of visual memory to large translations in short-term and online learning.
We construct the encoder following VGG \cite{Simonyan:2015ws} and concatenate 5 deconvolutional blocks \cite{long2015fully} for decoder.

\subsection{Short-term Learning}\label{sec:short-term}
As aforementioned, we normally only know the uninteresting scenes before a robotic mission is started. For known interesting objects, we prefer to use supervised object detectors.
Therefore, we expect that our unsupervised model can be trained \textit{incrementally} with negative labeled samples within \textit{several minutes}.
This is beneficial for learning environmental knowledge and quick robot deployment.

To this end, we propose the short-term learning architecture in \fref{fig:short-term-learning}. The memory module is inserted into the trained reconstruction model, in which all parameters are frozen.
For each sample, the output of the encoder is first written into the memory, then memory reading is taken as inputs of the decoder.
Intuitively, the images cannot be reconstructed well initially, as feature maps are not fully learned by the memory, and memory reading will be different from the encoding outputs. In this sense, we can inspect the reconstruction error to know whether the memory has learned to encode the training samples or not.

The memory leaning is much faster than back-propagation and has several advantages.
Recall that the gradient descent algorithms cannot be directly applied to neural networks for incremental learning, since all trainable parameters are changed during training, leading the model to be biased towards the augmented data (new negative labeled data), and forgetting the previously learned knowledge.
Although we can train the model on the entire data, which takes the learned parameters from long-term learning as an initialization, it is too computationally expensive and cannot meet the requirements for short-term learning.
Nevertheless, memory learning is able to solve this problem inherently.
One of the reasons is that the tangent operator in \eqref{eq:writing-vector} promotes writing sparsity, thus less memory cubes are affected, resulting in safer and faster incremental learning.

\subsection{Online Learning}\label{sec:online}

Online learning is one of the most important capabilities for a real-time visual inerestingness prediction system, as human feelings always keep changing according to one's environments and experiences.
Moreover, people tend to lose interests when repeatedly observing the same objects or exploring the same scenes, which is very common in a video stream from a mobile robot.
Therefore, we aim to establish such an online learning capability for real-time robotic systems, instead of selecting interesting frames after processing an entire sequence \cite{constantin2019computational}.
We design a few control variables that can be simply adjusted for different applications, \eg a hyper-parameter to control the rate of losing interests will be useful for objects search.
To this end, we propose an architecture for online learning in \fref{fig:online-learning}, in which only the frozen encoder and memory are involved.

In this stage, memory reading is performed before writing and the inputs are continuous image sequences (a video stream), which is different from short-term learning.
If unobserved scenes or objects appear suddenly, memory reading confidence will be lower than before, which can be treated as a new interest. Moreover, since the new scenes or objects are then written into the memory, their reading confidence level will become higher in the following images. Therefore, the model will learn to lose interests on repetitive scenes once the scene is remembered by the memory. In this sense, a visual interestingness is negative correlated with the memory reading confidence. In experiments, we adopt averaged cosine similarity over feature channel to approximate the reading confidence.

During online learning, a large translation often happens during robot exploration, hence an invariance to large translations introduced in \eqref{eq:reading} is able to further reduce memory consumption and improve the system robustness.

\section{Experiments}\label{sec:experiments}

\subsubsection{Evaluation Metric}~
Prior research typically only focused on the precision or recall rate and is not able to capture the online response of interestingness.
Therefore, we propose a new metric, \ie area under curve of online precision (AUC-OP) to evaluate one frame without using the information from its subsequent frames (no data leakage).
This metric is stricter and jointly consider online response, precision, and recall rate.
Intuitively, if $K$ frames of a sequence are labeled as interesting in the ground truth, an algorithm is perfect if the set of its top K interesting frames are the same with the ground truth.

Consider a sequence $I_{[1:N]}$, we take an interestingness prediction $p(I_t)$ as a true positive (interesting) if and only if $p(I_t)$ ranks in the top $K_{t,n}$ among a subsequence $p(I_{t-n+1})$, $p(I_{t-n+2})$ $\cdots$, $p(I_{t})$,  where $K_{t,n}$ is the number of interesting frames in the ground truth.
Note that the subsequence $I_{[t-n+1:t]}$ only contains frames before $I_t$, as data leakage is not allowed in the online performance.
Therefore, we may calculate an online precision score for length $n$ subsequences as
$s(n) = \nicefrac{\sum\text{TP}}{\left(\sum\text{TP}+\sum\text{FP}\right)}$, where TP and FP denote the number of true positives and false positives, respectively.
Since all true positives rank in the top $K_{t,n}$, this means that no false negative is allowed. Recall that a recall rate can be calculated as
$r=\nicefrac{\sum\text{TP}}{\left(\sum\text{TP}+\sum\text{FN}\right)}$, which means that the proposed online precision score $s(n)$ requires a $100\%$ recall rate.
For a better comparison, we often accept true positive predictions as ranking in the top $\delta \cdot K_{t,n}$, where $\delta \geq 1$.
Therefore, the overall performance of that jointly considers online performance, precision, and recall rate is the AUC of online precision $s(\frac{n}{N}, \delta)$ where $\frac{n}{N}\in(0,1]$, which considers all subsequence length as $n=[1:N]$. In practice, we often allow some false negatives and $\delta=2$ is recommended for most of exploration task.

\begin{table}[!t]
    \caption{The SubT dataset. ``Normal'' and ``Difficult'' means that the percentage of frames labeled as interesting by at least 1 subjects or 2 subjects, respectively.}
    \begin{center}
        \begin{tabular}{C{0.2\linewidth}C{0.08\linewidth}C{0.08\linewidth}C{0.08\linewidth}C{0.08\linewidth}C{0.08\linewidth}C{0.08\linewidth}C{0.08\linewidth}C{0.1\linewidth}}
            \toprule
            Video & \rom{1} & \rom{2} & \rom{3} & \rom{4} & \rom{5} & \rom{6} & \rom{7} & Overall\\
            \midrule
            Length (\minute) & 53.1  & 55.7  & 79.4  & 80.0 & 59.0 & 57.5 & 83.0 & 467.7 \\
            Normal (\%) & 11.11 & 15.07 & 9.37 & 17.51 & 24.52 & 22.77 & 11.04 & 15.14\\
            Difficult (\%) & 2.76 & 4.49  & 3.02 & 4.29 & 4.07 & 3.30  & 3.21 & 3.58\\
            \bottomrule
        \end{tabular}
    \end{center}
    \label{tab:subt}
\end{table}

\subsubsection{Dataset}~
To test the online performance on robotic systems for visual interesting scene prediction, we choose two datasets recorded by fully autonomous robots, \ie the SubT dataset \cite{subtdata} for unmanned ground vehicles (UGV) and the Drone Filming dataset \cite{wang2019improved} for unmanned aerial vehicles (UAV).

\begin{figure*}[!t]
    \centering
    \begin{subfigure}[t]{0.49\textwidth}
        \centering
        \includegraphics[width=1\linewidth]{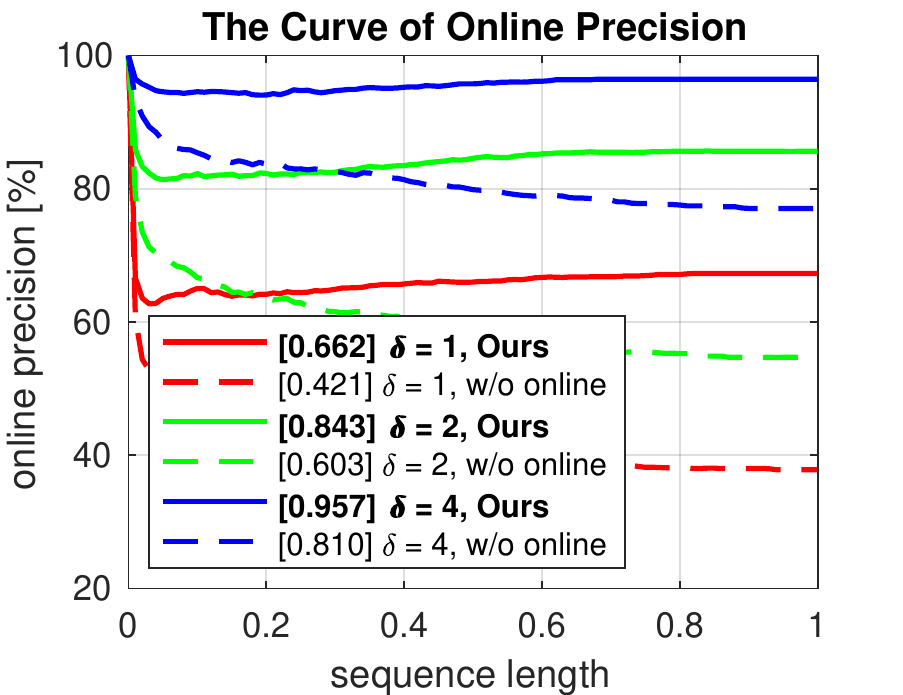}
        \caption{The Normal Category.}
        \label{fig:subt-union}
    \end{subfigure}
    \begin{subfigure}[t]{0.49\textwidth}
        \centering
        \includegraphics[width=1\linewidth]{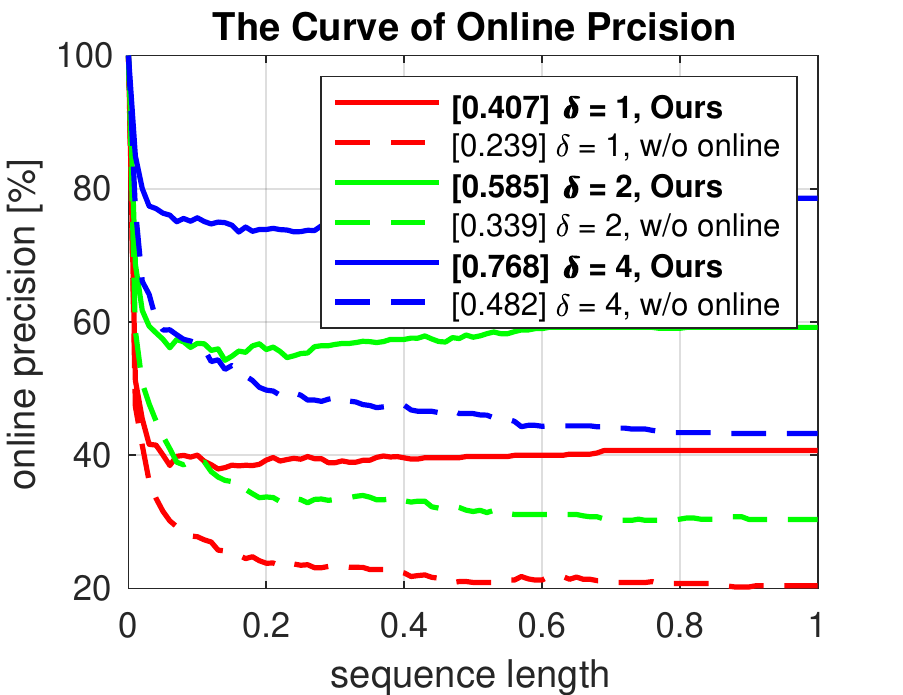}
        \caption{The Difficult Category.}
        \label{fig:subt-intersection}
    \end{subfigure}%
    \caption{The performance on SubT with and without (w/o) online learning.}
\end{figure*}

The SubT dataset is based on the DARPA Subterranean Challenge (SubT) Tunnel Circuit.
In this challenge, the competitors are expected to build robotic systems to autonomously search and explore the subterranean environments. The environments pose significant challenges, including a lack of lighting, lack of GPS and wireless communication, dripping water, thick smoke, and cluttered or irregularly shaped environments.
Each of the tunnels has a cumulative linear distance of 4-8 \kilo\meter.
The dataset listed in \tref{tab:subt} contains seven long videos ($1\hour$) recorded by two fully autonomous UGV from Team Explorer\footnote{\href{https://www.subt-explorer.com}{Team Explorer} won the first place at the DARPA SubT Tunnel Circuit.}.
Each sequence is evaluated by at least 3 persons.
It can be seen that the SubT dataset is very challenging, as human annotation varies a lot, \ie only 15\% and 3.6\% of the frames are labeled as interesting by at least 1 (normal category) and 2 subjects (difficult category), respectively. Some of the interesting scenes predicted by our algorithms are presented in \fref{fig:motivation}, in which we can see that our method predicts many interesting scenes correctly.

The Drone Filming dataset \cite{wang2019improved}  is recorded by quadcopters during autonomous aerial filming.
It also contains challenging environments, \eg intensive light changes, severe vibrations, and motion blur, \etc Different from other sources such as surveillance camera, robotic visual systems pose extra challenges due to fast background changes, limited computational resources, and unique and even dangerous operating environments in which human beings cannot get access to.

\begin{table}[!t]
    \caption{The comparison with the state-of-the-art method on AUC-OP.}
    \begin{subtable}{.5\linewidth}
        \centering
        \caption{The SubT Normal Category.}
        \label{tab:normal-compare}
        \begin{tabular}{C{0.35\linewidth}C{0.18\linewidth}C{0.18\linewidth}C{0.18\linewidth}}
            \toprule
            Methods & $\delta=1$  & $\delta=2$  & $\delta=3$  \\
            \midrule
            baseline \cite{liu2018future} & 0.622 & 0.798 & 0.904  \\
            ours & \textbf{0.662} & \textbf{0.842} & \textbf{0.923} \\
            \bottomrule
        \end{tabular}
    \end{subtable}%
    \begin{subtable}{.5\linewidth}
        \centering
        \caption{The SubT Difficult Category.}
        \label{tab:difficult-compare}
        \begin{tabular}{C{0.35\linewidth}C{0.18\linewidth}C{0.18\linewidth}C{0.18\linewidth}}
            \toprule
            Methods & $\delta=1$  & $\delta=2$  & $\delta=4$  \\
            \midrule
            baseline \cite{liu2018future} & 0.352 & 0.544 & 0.700  \\
            ours & \textbf{0.407} & \textbf{0.585} & \textbf{0.768}  \\
            \bottomrule
        \end{tabular}
    \end{subtable}
\end{table}

\subsubsection{Implementation}~
In all experiments in this section, a memory capacity of 1000 and mean square error (MSE) loss are adopted. The memory reading and writing rate are set as $\gamma_r=\gamma_w=5$.
Our algorithm is implemented using the PyTorch library \cite{paszke2017automatic} and conducted on a single Nvidia GPU of GeForce GTX 1080Ti.

\subsubsection{Efficiency}~
During long-term learning, we perform unsupervised training with the coco dataset \cite{lin2014microsoft}. It takes about 3 days running on single GPU. For short-term learning, our model takes about 10 minutes for learning 912 uninteresting images in the SubT dataset, which is feasible for deployment purpose of most practical applications. For online learning, it runs about 72.01\milli\second~per frame on single GPU, which is feasible for real-time\footnote{Real-time means processing images as fast as human brain, \ie  100\milli\second/frame \cite{potter1969recognition}.} robotic interestingness prediction.

\subsubsection{Performance}~
Online learning is able to remove many repetitive scenes thus it is able to reduce the number of false positives dramatically. The curve of online precision of our model for the normal category and difficult category are presented in \fref{fig:subt-union} and \fref{fig:subt-intersection}, respectively, where the overall AUC-OP is shown in the associated square brackets.
It can be seen that our model achieves an average of 20\% higher overall performance than the model without online learning, which verifies the importance of the proposed online learning.

\subsubsection{Comparison}~
To the best of our knowledge, robotic visual interestingness prediction is currently underexplored, and existing methods in saliency or anomaly detection have poor performance in this scenario.
In this section, we select the state-of-the-art method, frame prediction in \cite{liu2018future} as the baseline, which has very good generalization ability.
Basically, it introduces temporal constraint into the video prediction task to detect anomaly.
The overall performance of the AUC-OP of our method is presented in \tref{tab:normal-compare}  and \tref{tab:difficult-compare}, respectively. It can be seen that our method achieve an average of 4.0\%, 4.4\%, 1.9\% and 5.5\%, 4.1\%, 6.8\% higher overall accuracy in the two categories for $\delta=1,~2,~4$, respectively, which verifies its effectiveness.
We next present analysis to show the effects of the proposed writing sparsity, translational invariance, and short-term learning.

\mysubsec{Effect of Writing Sparsity}~
To show its effectiveness, we replace our proposed writing protocol with the one used in \cite{graves2014neural}, which is denoted as `without (w/o) sparsity' in the first row of \tref{tab:overall}. It can be seen that our model achieves about 20-30\% higher overall accuracy, which verifies the effectiveness of our method.

\mysubsec{Effect of Translational Invariance}~
Without the large translational invariance, the performance will drop a lot, as translational movement is very common in robotic applications.
As shown in the second row of \tref{tab:overall}, our model achieves about 20-30\% higher accuracy than the one w/o translational invariance.

\mysubsec{Effect of Short-term Learning}~
Short-term learning plays an important role for quick robot deployment. The performance can be largely improved if some uninteresting scenes are known before a mission. It can be seen in the fourth row of \tref{tab:overall} that our model achieves about 10-20\% higher accuracy than the one without short-term learning (w/o short-term).

\begin{table}[!t]
    \caption{The effects of the proposed modules on SubT (AUC-OP).}
    \label{tab:overall}
    \begin{center}
        \begin{tabular}{C{0.3\linewidth}C{0.1\linewidth}C{0.1\linewidth}C{0.1\linewidth}C{0.1\linewidth}C{0.1\linewidth}C{0.1\linewidth}}
            \toprule
            \multirow{2}{*}{Methods} & \multicolumn{3}{c}{Normal Category} & \multicolumn{3}{c}{Difficult Category} \\
            & $\delta=1$  & $\delta=2$  & $\delta=4$  & $\delta=1$ & $\delta=2$ & $\delta=4$\\
            \midrule
            w/o sparsity & 0.437 & 0.633  & 0.846 & 0.260 & 0.373 & 0.523 \\
            w/o invariance & 0.330 & 0.510  & 0.752 & 0.212 & 0.268 & 0.379 \\
            w/o short-term & 0.508 & 0.711 & 0.913 & 0.329 & 0.450 & 0.621 \\
            ours & \textbf{0.662} & \textbf{0.842} & \textbf{0.957} & \textbf{0.407} & \textbf{0.585} & \textbf{0.768} \\
            \bottomrule
        \end{tabular}
    \end{center}
    \label{tab:subt-comparison}
\end{table}

\section{Ablation Study}\label{sec:ablation}

In this section, we further test the proposed algorithm and aim to provide intuitive explanations for the influences of the proposed writing protocol, memory capacity, translational invariance, and capability of losing interests.
Following the ablation principle, all configurations are the same unless otherwise stated.

\begin{figure*}[!t]
    \centering
    \begin{subfigure}[t]{0.48\textwidth}
        \centering
        \includegraphics[width=1\linewidth]{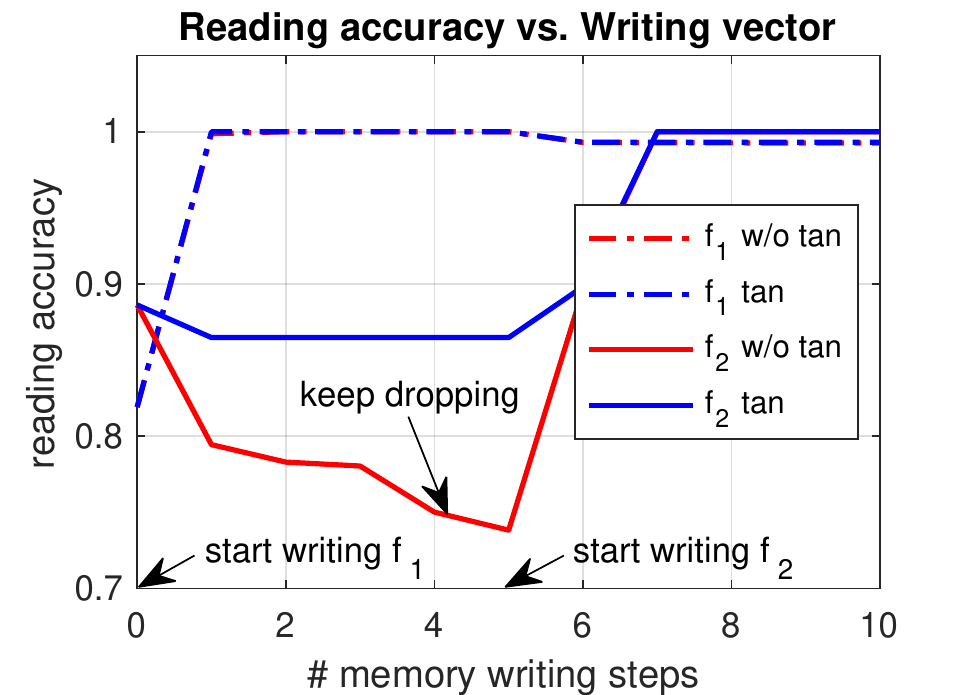}
        \caption{The influence of writing vector.}
        \label{fig:writing-vector}
    \end{subfigure}
    \hspace{0.01\textwidth}
    \begin{subfigure}[t]{0.48\textwidth}
        \centering
        \includegraphics[width=1\linewidth]{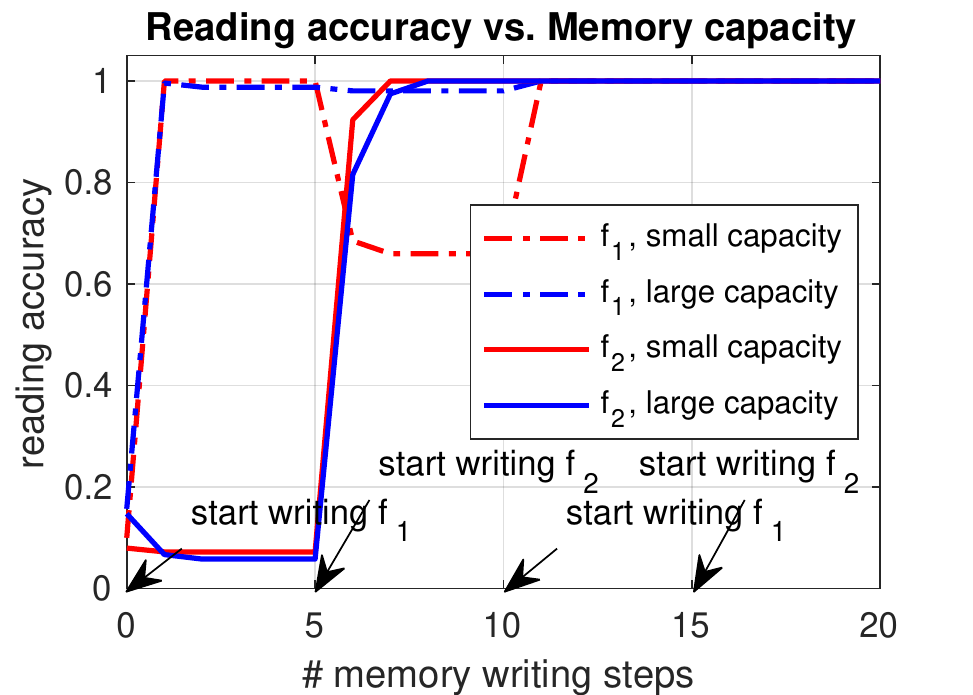}
        \caption{The influence of memory capacity.}
        \label{fig:memory-capacity}
    \end{subfigure}%
    \caption{The memory recall accuracy. (a) Writing vector with a tangent operator enables sparsity, thus less memory cubes are affected in learning. (b) Larger memory capacity leads to more computation but easy incremental learning.}
\end{figure*}

\subsection{Writing Protocol}\label{sec:writing-vector}

It has been pointed out that the memory learning is highly dependent on the writing vector in \eqref{eq:writing}, in which a tangent operator is introduced for writing sparsity.
This section explores this effect and compare with the writing vector in \eqref{eq:writing-vector-gamma} used in \cite{graves2014neural}. Note that memory defined in \cite{graves2014neural} is vectors, thus it is not invariant to large translation.
Following the ablation principle, we use the same 4-D memory structure and the reading protocol proposed in this paper.
\begin{equation}\label{eq:writing-vector-gamma}
\mathbf{w} = \softmax(\gamma \cdot D(\mathbf{x}, \mathbf{M})),
\end{equation}
where $\gamma$ is a parameter.
To show the writing performance, we write two random 3-D tensors into the memory, \ie $\mathbf{f}_1$ and $\mathbf{f}_2$, and compare their reading accuracy in terms of cosine similarity in \eqref{eq:reading-accuracy}.
\begin{equation}\label{eq:reading-accuracy}
S^c(\mathbf{r}, \mathbf{f}) = \frac{\sum(\mathbf{r}\odot \mathbf{f})}{\|\mathbf{r}\|_\mathbf{F}\cdot\|\mathbf{f}\|_\mathbf{F}},
\end{equation}
where $\mathbf{r}$ and $\mathbf{f}$ are the memory reading and writing tensors, respectively.
In experiments, we set $\gamma_w=\gamma=5$ and write both $\mathbf{f}_1$ and $\mathbf{f}_2$ 5 times continuously and show their reading accuracy in terms of number of writing in \fref{fig:writing-vector}. It can be seen that both memories are able to remember the random tensors after repeatedly writing. However, when writing a vector without a tangent operator, the accuracy of $\mathbf{f}_2$ keeps dropping even when $\mathbf{f}_1$ is learned, \ie $S^c(\mathbf{r}_1, \mathbf{f}_1)\approx1$. This is because all memory cubes are affected due to the non-sparse writing vector in \eqref{eq:writing-vector-gamma}.
This will be a severe issue when a robot keeps learning the same thing (observing the same scene), since the learned knowledge may be forgotten due to the non-sparse writing.
Nevertheless, our proposed writing vector with the tangent operator is able to map the weight of $\mathbf{f}_1$ to infinite  when $\mathbf{f}_1$ is learned, resulting in safer writing as only a few memory cubes are affected. This verifies the effectiveness of the proposed writing vector.
We notice that sparsity is also mentioned in \cite{zhao2011online,luo2017revisit,gong2019memorizing}, while it is designed for different objectives using different strategies. For instance,  \cite{gong2019memorizing} introduced a simple threshold and an entropy loss to promote sparsity for reducing reconstruction accuracy to detect anomaly.

\subsection{Memory Capacity}

This section explores the effects of memory capacity, \ie the number of memory cubes $c$, which is an important hyper-parameter for incremental learning.
To this end, we write two same random 3-D tensors $\mathbf{f}_1$ and $\mathbf{f}_2$ five times sequentially into two different memories in terms of the memory capacity $c$.
Their reading accuracy for the performance comparison is shown in \fref{fig:memory-capacity}.

As can be seen, both memories are able to learn random samples, while the accuracy of $\mathbf{f}_1$ drops a lot for smaller capacity when start to write $\mathbf{f}_2$, although it is remembered later when $\mathbf{f}_1$ is written again.
We observe similar phenomenon when the number of samples is around the same or larger than the memory capacity.
This means that a memory that has a small capacity quickly forgets old knowledge when learning new knowledge.
We can also leverage this property for model design, since uninteresting objects can become interesting in some cases.
This means that for larger capacity, reading accuracy is less affected by new knowledge, resulting in safer and easier incremental learning.

\begin{figure}[!t]
    \centering
    \includegraphics[width=0.75\linewidth]{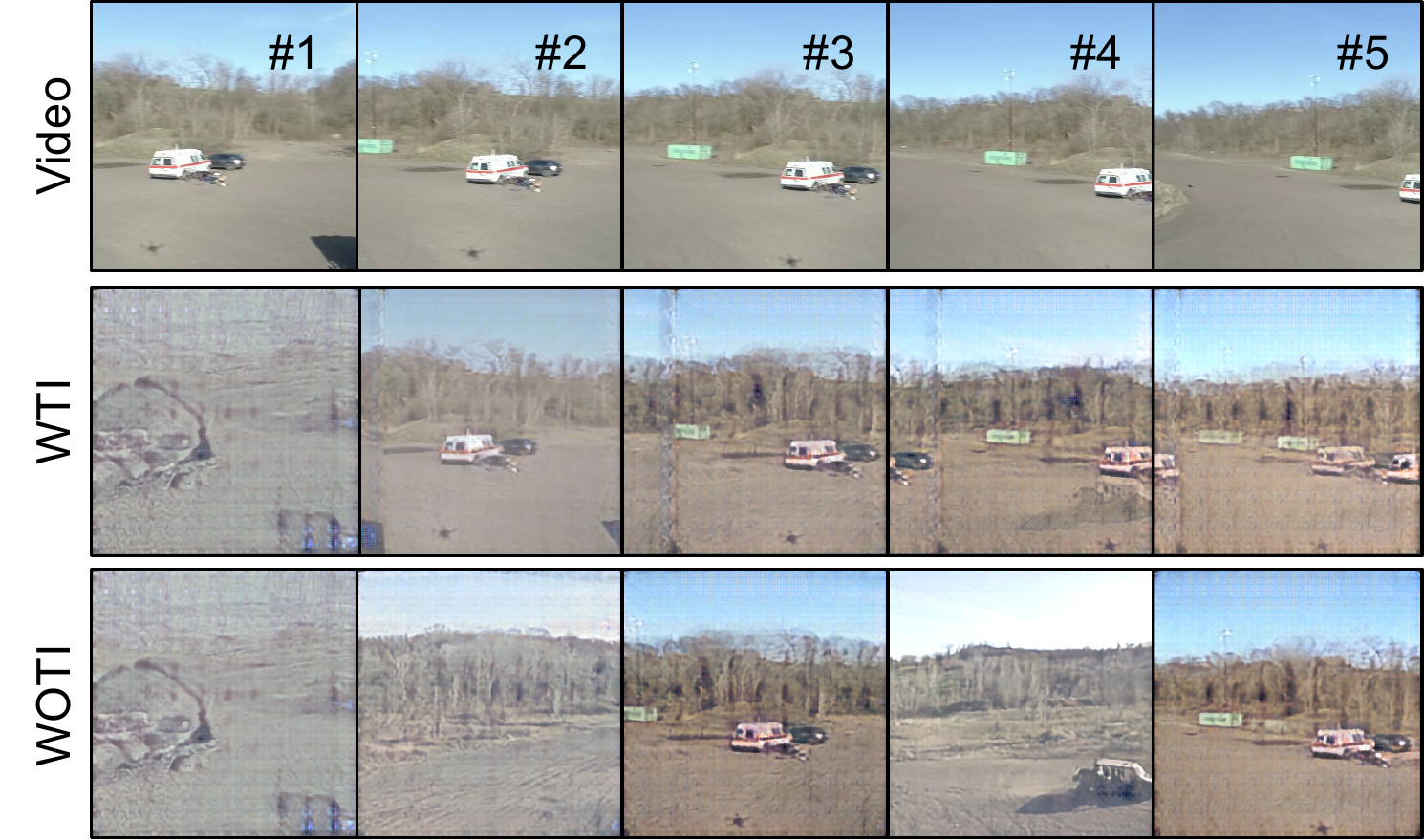}
    \caption{Memory reading with translational invariance (WTI) recall translated scenes better than without translation-invariance (WOTI). }
    \label{fig:translational-invariance}
\end{figure}

\subsection{Translational Invariance}\label{sec:translation-invariance}

Although CNN features are invariant to small translations \cite{wang2019kervolutional}, they still fail to recall memory when large translations occur.
To solve this problem, we introduce translational invariance by cross-correlation in \eqref{eq:cross-correlation}.
We next test it on the Drone Filming dataset \cite{wang2019improved} in \fref{fig:translational-invariance}. In this sequence an ambulance appears suddenly in the 1st frame and disappears in the 5th frame.
We construct two memory modules to learn this video based on the online learning strategy presented in \sref{sec:online}.
The first module adopts the cross-correlation similarity presented in \sref{sec:memory-reading} for memory reading (denote as WTI), while another one adopts the cosine similarity (WOTI).
It can be seen that both modules cannot recall the memory for the 1st frame, since the ambulance is not seen before.
However, the module WTI is able to recall the memory precisely in the subsequent frames, while the module WOTI quickly fails, although its reading is still meaningful, \eg the 2nd and 4th frames have correct patterns for sky, trees, and ground.
It can be seen that the recalled memory for the 3rd frame from module WTI is roughly a translated replica of the 2nd frame of the video (this also occurs at the 4th and 5th frame), which means that the module WTI correctly takes the 2nd frame as the most similar scene to the 3rd frame. This phenomenon verifies the translational invariance of our proposed reading protocol.

Note that there is a small translation for the 2nd frame from WTI.
This is because the invariance to small translations of CNN features, \ie  the features look the same for visual memory, although they appear with a small translational difference.
Therefore, our proposed cross-correlation similarity together with the CNN features contribute complete invariance of translation to memory recall.

\begin{figure*}[!t]
    \centering
    \includegraphics[width=1.0\linewidth]{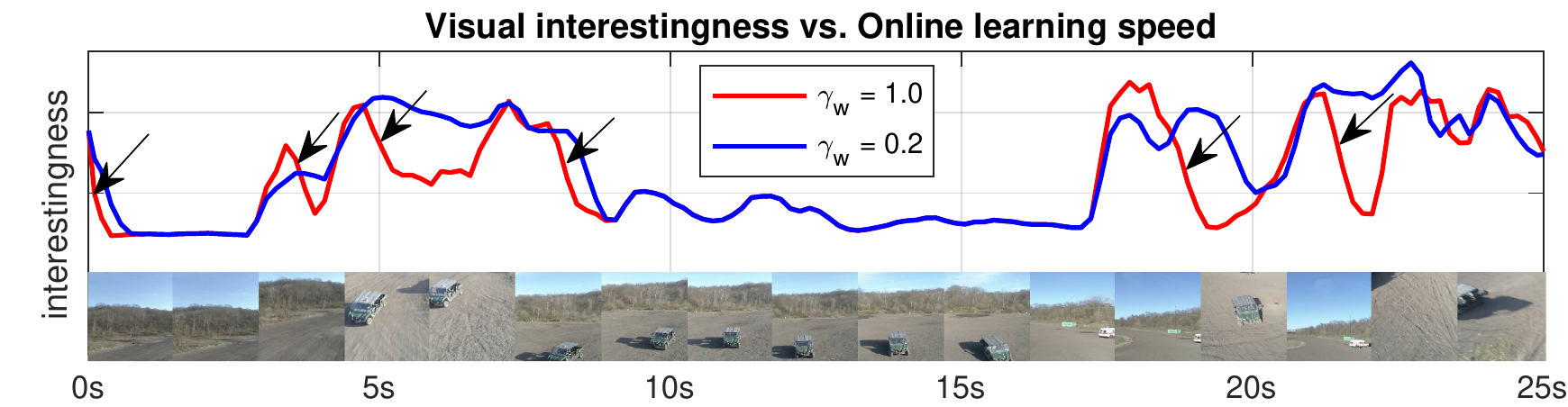}
    \caption{The visual interestingness with different writing rates for drone video footage \cite{wang2019improved}. As indicated by the arrows, a larger writing rate results in a faster loss of interest for new objects during online learning.}
    \label{fig:drone-filming}
\end{figure*}

\subsection{Losing interest}

To test the capability of losing interest of the algorithm, we perform a qualitative test on the Drone Filming dataset \cite{wang2019improved}.
The objects tracked in the videos, \eg cars or bikes, are relatively stable, while the background keeps changing due to the movement of the objects.
This makes it suitable for testing the capability of online learning.
One of the video clips is shown in \fref{fig:drone-filming}, where two different online learning speeds are adopted, \ie $\gamma_w = 1.0$ and $\gamma_w =0.2$.
It can be seen that the interestingness level of both settings become high when new objects or scenes appear, \ie both settings are able to detect novel objects.
However, the interestingness level with a larger writing rate always drops faster, meaning it is quicker to lose interest of the similar scenes.
This verifies our objective that a simple hyper-parameter can be adjusted for different missions.


\section{Conclusion}
In this paper, we developed an unsupervised online learning algorithm for visual robotic interestingness prediction.
We first proposed a novel translation-invariant 4-D visual memory, which can be trained without back-propagation.
To better fit for practical applications, we designed a three-stage learning architecture, \ie long-term, short-term, and online learning.
Concretely, the long-term learning stage is responsible for human-like life-time knowledge accumulation and trained on unlabeled data via back-propagation.
The short-term learning is responsible for learning environmental knowledge and trained via visual memory for quick robot deployment.
The online learning is responsible for environment adaption and leverage the visual memory to identify the interesting scenes.
The experiments show that, implemented on a single machine, our approach is able to learn online and find interesting scenes efficiently in real-world robotic tasks.

\section*{Acknowledgements}
This work was sponsored by ONR grant \seqsplit{\#N0014-19-1-2266}. The human subject survey was approved under \seqsplit{\#2019\_00000522}.



%
%
\bibliographystyle{splncs04}
\bibliography{egbib,papers,mypublication}

\clearpage
\appendix

\section{Interesting Scenes in SubT}
\begin{figure}[!h]
    \centering
    \includegraphics[width=0.95\linewidth]{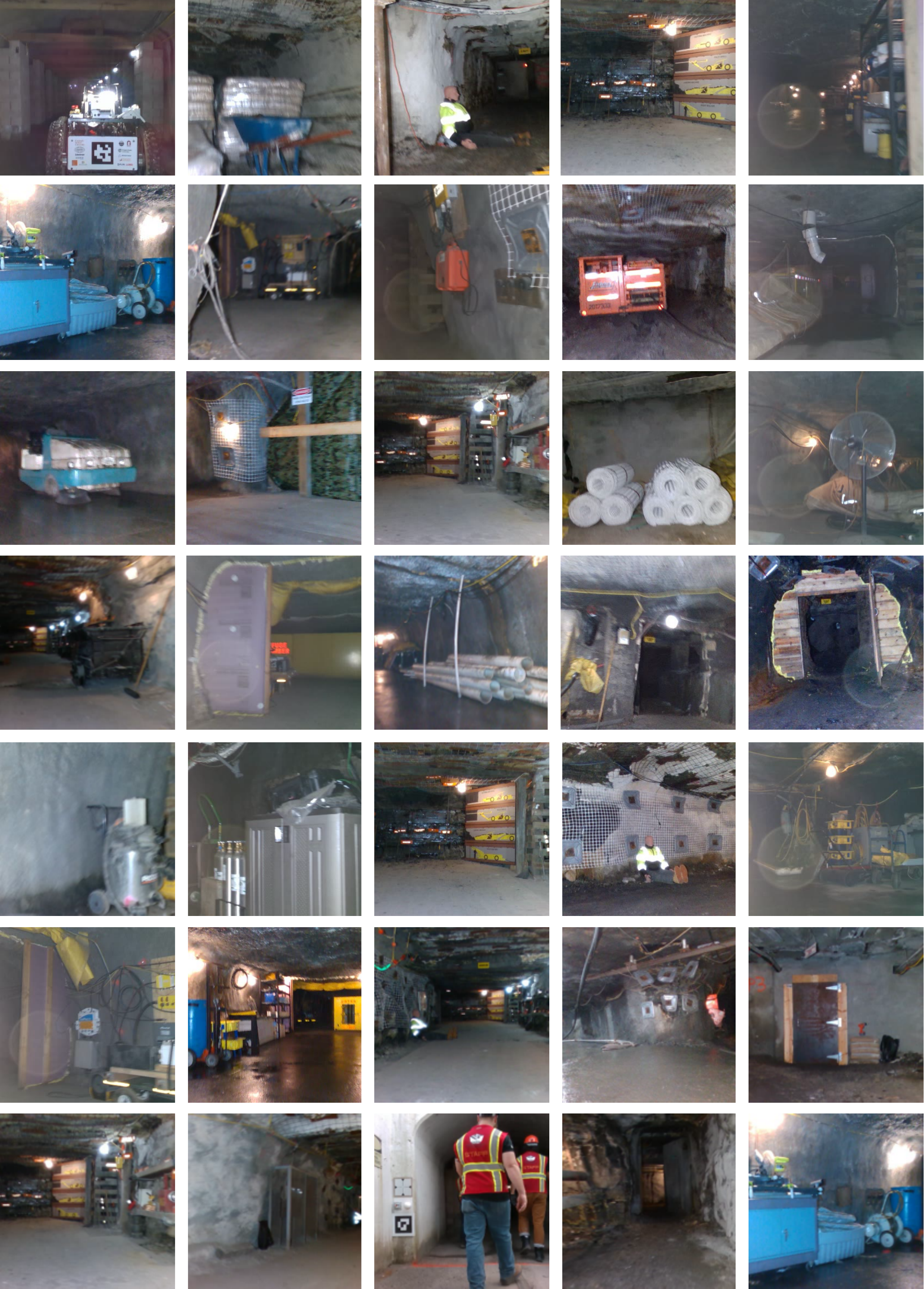}
    \caption{The examples of the detected interesting scenes.}
    \label{fig:interestingness}
\end{figure}

\fref{fig:interestingness} shows the detected interesting scenes from SubT dataset.
It can be seen that many interesting scenes, e.g., doors and intersections, are detected.

\end{document}